\documentclass{article}
\usepackage{amsmath} 

\usepackage[preprint]{neurips_2026}

\usepackage[utf8]{inputenc} 
\usepackage[T1]{fontenc}    
\usepackage{hyperref}       
\usepackage{url}            
\usepackage{booktabs}       
\usepackage{amsfonts}       
\usepackage{nicefrac}       
\usepackage{microtype}      
\usepackage{xcolor}         
\usepackage{graphicx}
\usepackage{tabularx}
\usepackage{subcaption} 

\usepackage[framemethod=TikZ]{mdframed}
\usepackage{helvet} 

\definecolor{headerblue}{HTML}{DCE6FF}   
\definecolor{envblue}{HTML}{4A7BB0}      
\definecolor{tealtext}{HTML}{4A908A}     

\mdfdefinestyle{cyclereactbox}{%
    linecolor=black,
    linewidth=1pt,
    roundcorner=10pt,
    backgroundcolor=white,
    innerleftmargin=14pt,
    innerrightmargin=14pt,
    innertopmargin=10pt,
    innerbottommargin=12pt,
    frametitlefont=\bfseries,
    frametitlebackgroundcolor=headerblue,
    frametitlerule=true,
    frametitlerulecolor=black,
    frametitlerulewidth=1.4pt,
}
\title{The Capability Manifold and ML Scaling Laws}

\author{%
  Syed~Ali~Raza~Zaidi\ \\
  SPRINT Lab\\
  University of Leeds\\
  Leeds, LS2 9JT, UK. \\
  \texttt{s.a.zaidi@leeds.ac.uk} \\
  \And
  Maryam Hafeez \\
  SPRINT Lab \\
  University of Leeds \\
  Leeds, LS2 9JT, UK.\\
  \texttt{m.hafeez@leeds.ac.uk} \\
}

\begin{document}

\maketitle

\begin{abstract}
 Existing machine learning (ML) scaling laws relate predictive loss to compute, model parameters, and data. However, as models are increasingly deployed through agentic harnesses, loss alone is insufficient to characterize downstream performance: models with similar loss can exhibit different capabilities in reasoning, retrieval, planning, and adaptation. Yet, no unified framework connects such capabilities to the coupled resources available across the ML lifecycle. We bridge this gap by introducing a capability manifold, a multidimensional framework mapping downstream capabilities to pre-training, post-training, and test-time resources through bounded scaling functions. Analytical Jacobians quantify capability sensitivity to resource changes and interactions. As an initial application, we embed Kaplan- and Chinchilla-type scaling laws and test-time compute within the framework, demonstrating how existing scaling relationships can be unified as trajectories on a common capability manifold.
\end{abstract}

\section{Introduction}
\subsection{Motivation}
Developing rigorous theoretical foundations to explain the \textbf{learning mechanics} of deep learning (DL) remains a key objective in machine learning (ML) research. \textbf{Scaling laws} established on the basis empirical evidence are an important component that underpins the emerging theory of learning mechanics \cite{bordelon2026theory}. Scaling laws were first investigated \cite{kaplan2020scaling}\cite{hoffmann2022training} to establish a concrete relationship between  model size, training data and computation and \textbf{predictive loss}. The goal was to understand optimal choices in terms of input parameters ($N$), training data ($D$), and employed compute ($C$) establishing how these resource should scale to minimise the predictive loss. Subsequent research works\cite{snell2024scaling}\cite{zhang2024scaling} have shown that this scaling behaviour extends beyond pre-training to fine-tuning, preference optimisation and test-time computation. Scaling laws therefore provide a common language for relating the resources invested in learning to the resulting performance of increasingly capable models.
\\
However, with the evolution of these models being deployed in \textbf{agentic AI} harness, minimisation of loss is not the ultimate object of interest. We argue that the relations expressed through scalar observables e.g. pre-training or task loss should no longer be the core metric as modern foundation models exhibit a collection of distinct \textbf{functional capabilities}. Predicting downstream capabilities from scale has consequently remained substantially less direct than predicting loss \cite{schaeffer2024has}. We therefore, argue that one needs a unified framework to understand model \textbf{capabilities}. Moreover, when these capabilities are jointly considered they form what we call \textbf{capability manifold}. There are several reasons for developing this unified framework. To summarise a few:
\begin{enumerate}
    \item The distinction between studying capabilities and loss becomes important as scaling extends across various model lifecycle stages. Loss can be improved through different mechanisms, including pre-training, post-training and inference-time computation, but these mechanisms need not produce the same changes in individual capabilities.
    \item Similarly, operations such as model compression and knowledge distillation change the available resource configuration while attempting to preserve selected capabilities, whereas safety interventions may deliberately modify particular behavioural dimensions without simply minimising loss. More generally, different capabilities may scale at different rates, saturate at different points, or depend on different combinations of resources. Thus, the relevant question is not only how much a model scales, but which resources are scaled, at which stage, and along which capability axes.
\end{enumerate}
We therefore require a framework that explicitly connects resources, scaling mechanisms and multidimensional capabilities rather than treating loss as the sole endpoint of scaling.
\subsection{Capability Aware Design}
At high-level the capability manifold can be mathematically defined as: 
\begin{equation}
    \mathcal{M}_C=\{\mathbf{C}(\mathbf{X})\}, \textrm{ with } \mathbf{C}(\mathbf{X}) = [C_1 (\mathbf{X}),C_2(\mathbf{X}),...,C_M(\mathbf{X})]
\end{equation}

where $\mathbf{X}=[N,D,C,...]$ is input resource vector, $C_i$ is $i^{\textrm{th}}$ capability, and, $\mathbf{C}(\mathbf{X})$ is the capability vector. The manifold formulation changes several practical model-design problems. Rather than optimising a scalar loss, one can seek the minimum-cost resource configuration satisfying a target capability vector,
\begin{equation}
    X^* = \arg\min_X \text{Cost}(X) \quad \text{s.t.} \quad C(X) \ge C^*
\end{equation}
This formulation naturally encompasses several settings. For knowledge distillation, the objective can be capability preservation under reduced model resources; for safety, it can express improvements in selected behavioural dimensions subject to constraints on other capabilities; and for inference-time scaling, it can determine when additional reasoning or search is preferable to increasing pre-training resources. The Jacobian of capability with respect to resource further provides a local measure of these trade-offs, while the global manifold captures nonlinear transitions and saturation.

\subsection{Contributions \& Organisation}
To that end, this paper makes three core contributions:
\begin{enumerate}
    \item  We introduce the capability manifold, representing model behaviour as a multidimensional capability vector whose coordinates can depend differently on pre-training, post-training and test-time resources. The capability definitions are operational and provide a common coordinate system rather than a prescription for universal benchmarks.
    \item We formulate heterogeneous scaling laws through bounded resource transformations, capability-specific mappings and cross-resource interactions, and derive the corresponding capability Jacobian. This provides a local measure of how changes in each resource affect individual capabilities and their joint trajectory.
    \item We instantiate the framework using established pre-training scaling relationships, including Kaplan and Chinchilla, together with test-time scaling and emergence, demonstrating how alternative resource allocations can produce distinct capability trajectories despite comparable aggregate scaling.

\end{enumerate}
The remainder of the paper is organised as follows. Section 2 reviews scaling relationships across pre-training, post-training, test-time inference and emergence. Section 3 introduces the capability vector and mathematical manifold formulation, followed by an application illustrating capability trajectories under alternative scaling strategies.
\section{AI \& ML Scaling Laws}
\label{scal_law}
We start by revisiting the fundamental scaling laws for AI \& ML established in the literature. Most of the discussed laws are established on the basis of \emph{empirical evidence} and can be understood through the lens of the life-cycle where scaling is experienced. 
\\
During the \textbf{pre-training stage}, the focus is on understanding the relationship between the resources (e.g., parameters, data, and, compute) and the model performance. The scaling law essentially characterise rate of performance improvement with the increase in the resources. \textbf{Post-training} scaling is geared to understand how pre-trained model can be improved through mechanisms such as supervised fine-tuning, preference optimisation, and reinforcement learning (RL), where additional resources (including data, feedback, and compute for optimisation) is utilised to improve the task related performance aspect.\textbf{Test-time scaling} shifts scaling variables from the construction of model to its deployment and utilisation. In essence, additional inference time compute, context, external tools, reasoning traces or agentic interactions can be used to improve the baseline model output. With reasoning coming into picture, the performance improvement can have temporal patterns, whereby the performance does not improve monotonically with increase in resources over time. However, a \emph{phase-transition} may exist whereby the improvement occurs as emergent behaviour when resources cross a certain threshold over time. We call scaling laws associated with such behaviour as \textbf{emergent behaviour} scaling laws. In summary, our subsequent discussion is organised in accordance with these discussed life-cycle stages.

\subsection{Pre-training Scaling Laws}
Pre-training scaling laws are geared towards quantification of the change in the performance with an increase in resources utilised to construct the model. Existing studies primarily consider three coupled resources: i) model size, measured by the number of trainable parameters $N$; ii) training data, measured by the number of tokens $D$; and iii) training compute,typically denoted by $C$ (measured in FLOPs or some derivative scale e.g. peta-flop (PF)-days). The fundamental question, we want to address is how does the training loss $L$ scales with model size $N$, dataset size $D$, and compute $C$. Consequently, a useful general formulation is:
\begin{equation}
L = f(N,D,C),
\label{preEq}
\end{equation}
where $L$ is usually cross-entropy or negative log-likelihood evaluated on held-out data.\\ 
A central empirical observation from the literature is that, over substantial ranges of model and training scale, loss often follows an approximate \textbf{power-law relationship} with each resource when the other resources are sufficiently controlled. Thus when holding two of the three factors constant, Eq. \ref{preEq} can be written in form:
\begin{equation}
L = L_f+ aX^{-\alpha},
\label{preEqg}
\end{equation}
with $a$ being some constant, $L_f$ being the asymptotic loss-floor, $a$ being a parameter dependent constant and $X\in\{C,D,N\}$ being the resource being varied.
\subsubsection{Parameter Scaling Law}

\citet{kaplan2020scaling} in  studied empirical scaling laws for language model performance on the cross-entropy loss $L$. Demonstrating that, for the models trained on sufficiently large dataset:
\begin{equation}
L(N) = \left(\frac{N_o}{N}\right)^{-\alpha_N},
\label{paraEq}
\end{equation}
where, $\alpha_n\sim 0.076$, and $N_o \sim 8.8 \times 10^{13}$ non-embedding parameters. Essentially, loss exhibits remarkably smooth power-law behaviour over a wide range of model sizes and training setups. Empirical results are derived on the basis of experiments covering models ranging from relatively small networks to billions of parameters. The results demonstrate that when models were sufficiently trained, increasing model size produced predictable reductions in loss. The exponent $\alpha_N$ is not universal and is found from the fit for the empirical experiments. The shallow exponents implies that the modest loss gain requires significantly higher increase in the parameters $N$. For instance, increasing the model size by $10\times$ yields $16\%$ reduction in loss, as shown below:
\begin{equation}
\bar{L}(N) = L(N)\times10^{-0.076}=0.84L(N).
\end{equation}
Another caveat here is that a reported scaling law is only valid when the other resources are sufficiently large. Moreover, the parameter size $N$ is not same as the architecture (or shape). In particular, for the Transformer architecture:
\begin{equation}
N = O(N_{L} d^2),
\end{equation}
where, $N_L$ is number of Layers and $d$ is model dimension. In \citet{kaplan2020scaling} it was shown that for their investigations model size $N$ was a much stronger predictor of loss than the precise allocation between depth $L$ and $d$ width. However, this again does not hold universally, as two models with $N_1=N_2$ but $(L_1,d_1)\neq(L_2,d_2)$ can have different loss scaling depending on the architecture. Subsequent work on efficient architectures, mixture-of-experts models etc. has practically validate this further. For the scope of our discussion, it will suffice to consider that $L\propto N^{-\alpha_N}$. 
\subsubsection{Data \& Compute Scaling Laws}

For investigating the impact of parameter size, we considered other resource dimension fixed. However, other important design questions are: i) what happens when model is exposed to increasing amount of data $D$; and ii) how does the model performance evolves as the amount of computation used to develop it increases? \\
Both these questions were answered by \citet{kaplan2020scaling}, showing that the loss follows power-law with respect to $D$ and $C$, when other parameters are constant. Mathematically,

\begin{equation}
L(D) = \left(\frac{D_o}{D}\right)^{-\alpha_D}, \alpha_D \sim 0.095, D_o \sim 5.4 \times 10^{13}.
\label{paraEqD}
\end{equation}
\begin{equation}
L(C) = \left(\frac{C^{min}_{c}}{C_{min}}\right)^{-\alpha^{min}_{c}}, \alpha^{min}_c \sim 0.050, C^{min}_c \sim 3.1 \times 10^{8}.
\label{paraEqD}
\end{equation}
\begin{figure}[htbp]
  \begin{subfigure}[htbp]{0.48\textwidth}
        \centering
        \includegraphics[width=\textwidth]{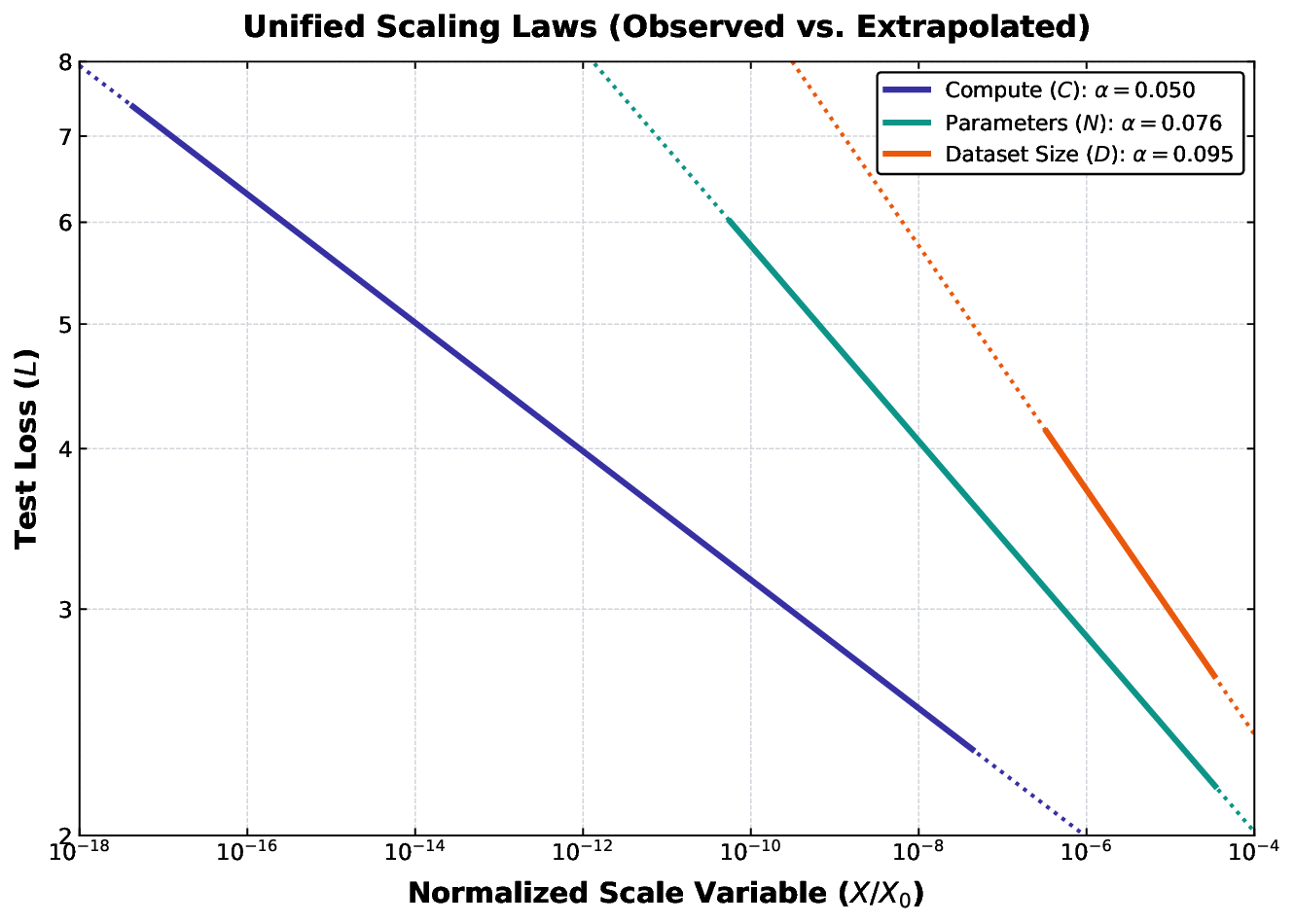} 
        \caption{First sub-figure caption.}
        \label{fig:scaling_laws}
    \end{subfigure}
    \hfill 
    \begin{subfigure}[htbp]{0.48\textwidth}
        \centering
        \includegraphics[width=\textwidth]{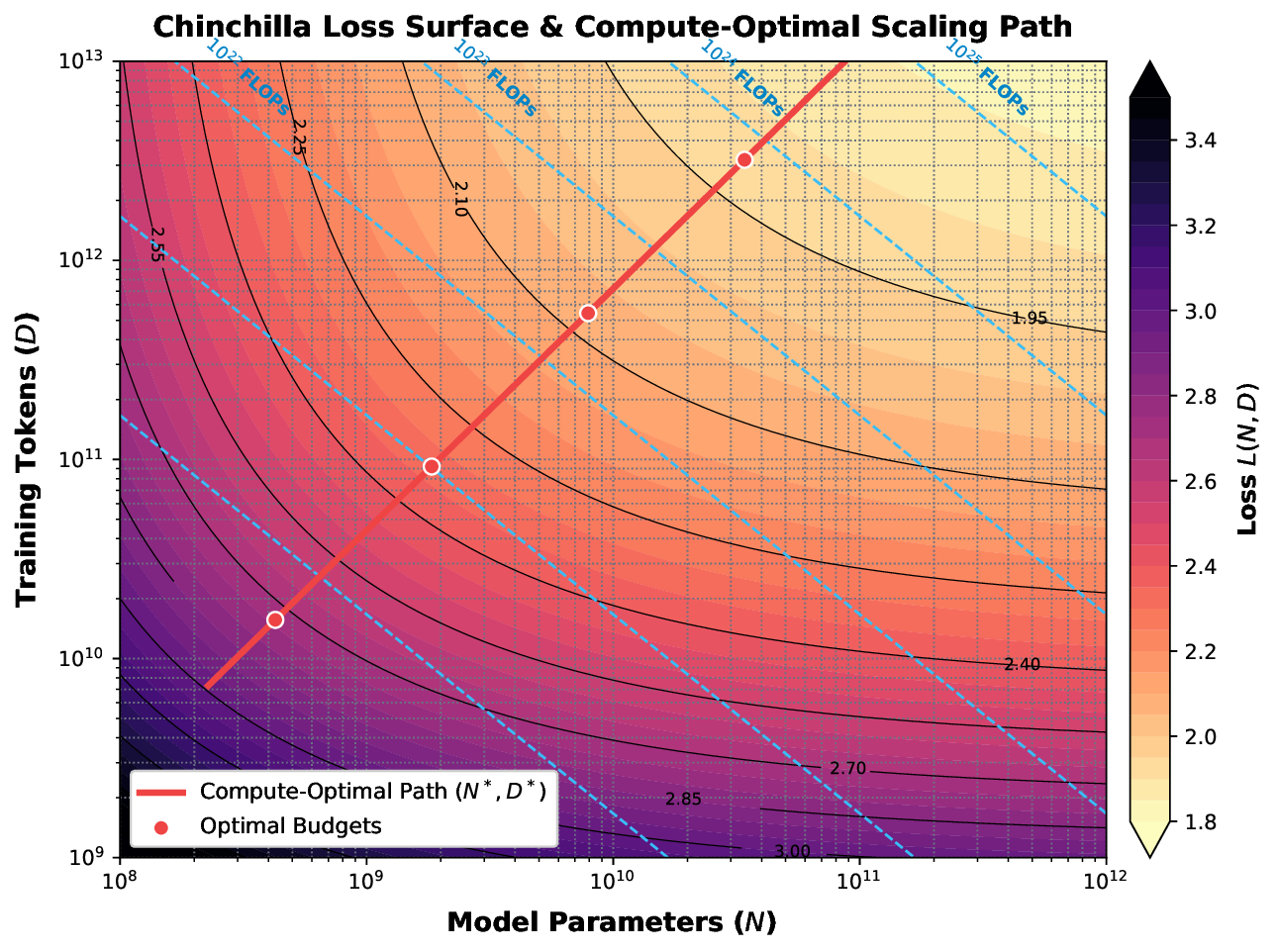} 
        \caption{Second sub-figure caption.}
        \label{fig:optimal_path}
    \end{subfigure}
    
    \caption{Main overall caption explaining both sub-figures.}
    \label{fig:main_figure}
\end{figure}
These relations were shown to hold across eight orders of magnitude in $C_{min}$, two orders of magnitude in $D$. Fig. \ref{fig:scaling_laws} plots all three power-law scalings on the normalised resource axis ($X/X_o \in \{C_{min}/C^{c}_{min}, D/D_o, N/N_o\}$). The solid line shows the range corresponding to \citet{kaplan2020scaling} and the dotted line shows the extrapolation of theoretical result. It is obvious that the slope of power-law relationship differs across the resources. This is governed by the difference in the exponents. In fact, \citet{kaplan2020scaling} showed that the loss function when expressed jointly over $N$ and $D$,depends predictably on the ratio $N^{0.74}/D$. As a result, the model size needs to increase faster than the data to avoid over-fitting. Notice that, the compute is not independent of $N$ and $D$ \cite{weng2026scaling}. In fact, $C\sim kND$, for the language models, and therefore it was shown that the optimal choice of $N_{opt}\propto C^{0.73}$ and $D_{opt}\propto C^{0.27}$. Kaplan's conclusion that the model size should increase more than data was revisited in \citet{hoffmann2022training}. In \cite{hoffmann2022training} it was shown that the hypothesis of scaling up the model size more than the data is not true and consequently:

\begin{equation}
C_{opt} \sim N^{0.5} \textrm{ and } C_{opt}\sim D^{0.5},
\label{paraEqC}
\end{equation}
So both the data and the model should simultaneously scale similarly to reduce the cross-entropy loss. Both papers show that it is not necessary to train the models to minimise the loss to absolute minimum. Chinchilla's law \cite{hoffmann2022training} as they are known provide general formulation for the parameter scaling laws for pre-training. Therefore, for the purpose of this paper, a combined scaling law in terms of the input parameters and the data set size can be stated as follows:
\begin{mdframed}[style=cyclereactbox, frametitle={Compute-Optimal Scaling Law}]
The cross-entropy loss in pre-training scale as:
\begin{equation}
L(N,D,C) = L_f + \frac{A}{N^\alpha} + \frac{B}{D^\beta}, \textrm{ with } C\propto ND
\label{paraEqFinalPre}
\end{equation}
and is compute optimal if:
\begin{equation}
\left(N_{opt} ,D_{opt}\right)  = \arg \min_{ (N,D)} L(N,D) \textrm{ s.t. } C=KND, 
\end{equation}
where $A,B, K$ are constants, $\alpha,\beta$ exponents are empirically found to be approximately $0.5$ and $L_f$ is the loss floor. 
\end{mdframed}
In Fig. \ref{fig:optimal_path} we plot Eq. \ref{paraEqFinalPre} then then the compute optimal configuration can be viewed as optimal path through the loss manifold, as shown in the figure. Interested readers are directed to \cite{weng2026scaling} and \cite{pearce2406reconciling} for detailed discussion on reconciliation of Chinchilla and Kaplan scaling laws. \\
It was shown in \cite{henighan2020scaling} that the scaling law of the form presented in Eq. \ref{paraEqFinalPre} holds beyond language model for four domains: generative image modelling, video modelling, multimodal image to text models, and mathematical problem solving. Nevertheless, one important point to notice is that pre-training loss is substantially more predictable from size than downstream capability\cite{schaeffer2024has}, and models with similar scale/loss can exhibit different downstream capability profiles. We will touch upon this in subsequent discussion.

\subsection{Post-training Scaling Laws}
While pre-training scaling laws are relatively well accepted and mature, post-training is fundamentally different. Post-training begins with a pre-trained model whose parameters encode the baseline knowledge and tries to modify the behaviour of such model using small amount of instructions, task, preference or reward-related additional data. The post-training process can therefore be viewed as a transformation: $\theta \rightarrow \theta_{PT}$ whereby $\theta$ are the model parameters. Essentially, the goal of post-training process is to minimise a performance metric $L_{PT}$ given $N_{FT}$ the fine-tuning data, $N_{pref}$ is preference data, $H$ denotes human or preference feedback, $C_{PT}$ is post-training compute, $N$ is pretrained model size, and $S$ may represent the number of optimisation steps. Mathematically,
\begin{equation}
L_{PT} = f(N,D,C,N_{FT},N_{pref},H,C_{PT},S).
\label{postEq}
\end{equation}
Instead of concrete empirical laws as exists in the case of pre-training, for post-training we have rather families for scaling as there is no single loss metric and post-training objectives may differ.
\subsubsection{Fine-tuning Scaling}
Supervised fine-tuning uses additional data-set as examples to align the model outcome:
\begin{equation}
D_{FT} = \{x_i,y_i\}^{N_{FT}}_{i=1}
\label{postDFT}
\end{equation}
The loss-function can be written as: 
\begin{equation}
L_{SFT} (N_{FT}) = -\frac{1}{N_{FT}} \sum_{i=1}^{N_{FT}}\log(p_\theta (y_i|x_i)), 
\label{posLSFT}
\end{equation}
In \citet{zhang2024scaling}, the authors investigated  how different scaling factors, including  model size, pre-training data size, new finetuning parameter size and finetuning data size, affect the finetuning performance. The authors considered both full fine-tuning as well as reduced parameter technique such as low rank adaptation  (LoRA). The finding is summarised below:  
\begin{mdframed}[style=cyclereactbox, frametitle={Multiplicative SFT Scaling}]
The cross-entropy loss in SFT scale as:
\begin{equation}
L_{SFT}(N_{FT},X) = L_f + \frac{A}{N_{FT}^\alpha} \times \frac{1}{X^\beta},
\label{paraEqFinalPost}
\end{equation}
where $A$ is data specific constant, $\alpha,\beta$ exponents and $L_f$ is the loss floor. Here $X$ are the other resources. As can be seen the law here is multiplicative, i.e., the resources and the methodology employed jointly impact the scaling. It was shown that scaling the pre-training model benefits fine-tuning more than the pre-training data. The scaling property for LLM finetuning is highly task- and data-dependent, making the selection of optimal finetuning method for a downstream task non-trivial.
\end{mdframed}
When considering the parameter efficient finetuning like LoRA, the size of parameters adapted is important as the pre-training model size is no longer is appropriate representative for loss-scaling.
\subsubsection{Preference and Reinforcement Learning with Human Feedback (RLHF) Scaling}
RLHF scaling  how alignment performance changes with preference data, model scale, and RL compute. A useful phenomenological form is
\begin{equation}
R_{RLHF}(H) = R_f + A H^{-\alpha},
\label{paraRLHF}
\end{equation}
where $H$ is function of preference feedback and human input. However, unlike pre-training loss scaling, this is not yet a universally established law. Empirical studies report power-law behaviour alongside saturation and regime dependence. Preference data therefore acts as a distinct scaling resource, with its quality and informativeness potentially as important as its quantity. 
\subsection{Test-time Scaling}
Test-time compute provides additional context during the \textbf{inferencing} phase. Empirical results in literature demonstrate scaling with demonstrations, reasoning compute, repeated sampling and search, but the functional form is highly dependent on the task and inference algorithm. We summarise these in Table \ref{tab:test_time_scaling}.
\begin{table*}[t]
\centering
\caption{Empirical scaling relationships in test-time inference.}
\label{tab:test_time_scaling}
\begin{tabular}{p{2cm} p{4.0cm} p{6.2cm} }
\hline
\textbf{Scaling} &
\textbf{Resource \& relationship} &
\textbf{Observation} \\
\hline

In-context learning &
Number of demonstrations $k$ is related to $L\propto k^{-\beta}$  &
Performance can improve with more demonstrations, but is strongly dependenton demonstration quality. 

\cite{bordelon2026theory}\\
\hline

Reasoning compute &
Reasoning tokens / inference compute $T$, it can be shown that accuracy $A\propto a \log T+b$ with $a,b$ being constants. &
Increasing inference-time reasoning compute can substantially improve
reasoning performance, generally with diminishing returns. 
\cite{snell2024scaling} \\

\hline

Repeated sampling &
Number of generated solutions $N$, then probability of success is $p_{suc} = 1-(1-p)^N$ &
Best-of-$N$, majority voting and related sampling methods improve success
probability as additional samples are generated.\cite{snell2024scaling} 
 \\

\hline

Agent count &
Number of agents $n_{\mathrm{agents}}$ &
Multi-agent scaling can improve performance on some tasks, but coordination overhead and correlated errors need to be tackled.
 \\

\hline

Tool use / horizon &
Tool calls $n_{\mathrm{tools}}$ and interaction horizon $G$ &
Longer agentic trajectories can increase task-solving capability. 
\\

\hline
\end{tabular}
\end{table*}

\subsection {Emergence and Phase Transition Scaling}
Emergent and phase-transition scaling relates to capabilities of a language model which does not exists in a smaller model but appears when for instance model size exceeds a certain threshold. In classical sense, there exists a phase-transition threshold which determines whether the models loss on a certain task can be minimised. \citet{wei2022emergent} first demonstrated that such behaviour exists. More recently, in \cite{wu2025u} showed an alternative way of inspecting the phase-transition threshold. Mathematically, emergence of capability can be described as:
\begin{equation}
C(N)=\frac{1}{1+\exp\big(-k\big(N-N_c \big) \big) },
\label{eqEmerge}
\end{equation}
where $N_c$ is the critical threshold.
To summarise the discussion thus far, we have revisited the mathematical principles which govern the learning mechanics which are known thus far. In what follows, we want to map these to the downstream capability measures.

\section{Capability Manifold}
In the previous section, we focused on scaling laws that describe the behaviour of models through a small number of scalar observables, most notably pre-training loss, as a function of model size, data and compute. However, we are fundamentally interested in model capability and not necessarily the reduction in loss. To that extent, a complete framework for model training should relate capabilities to underlying pre-training, post-training, test-time and emergent scaling laws. Then the multi-dimensional capability vector associated with the model lies on, what we call \textbf{capability manifold}. In the following discussion, we aim to formalise the notion of capability manifold. Our hope is that in future ML research, the community will be able to use this framework to provide unified comparison of model capabilities.

\begin{table*}[t]
\centering
\caption{Definitions of capabilities and their resource--capability relationships.}
\label{tab:capability_definitions}
\small
\renewcommand{\arraystretch}{1.35}
\begin{tabularx}{\textwidth}{@{} l >{\hsize=1.5\hsize}X >{\hsize=0.5\hsize}X @{}}
\toprule
\textbf{Capability} &
\textbf{Operational definition} &
\textbf{Resource ($X$) $C_i=f(X)$} \\
\midrule

\textbf{Knowledge} ($C_K$) &
Accuracy in retrieving and applying relevant knowledge,
$C_K=\frac{N_{\text{correct}}}{N_{\text{queries}}}$ &
$N, D, C, S$
\\

\textbf{Reasoning} ($C_R$) &
Success on tasks requiring multi-step inference or problem solving,
$C_R=\frac{N_{\text{correct}}}{N_{\text{tasks}}}$ &
$N, D, C, S$,\newline
$D_{\text{FT}}, D_{\text{pref}}, C_{\text{PT}}, k, T$
\\

\textbf{Planning} ($C_P$) &
Probability that an action sequence $\pi$ achieves the specified goal,
$C_P=\Pr(\text{goal achieved}\mid\pi)$ &
$S, T, H, B$
\\

\textbf{Tool use} ($C_T$) &
Ability to select and correctly invoke external tools,
$C_T=\frac{N_{\text{successful}}}{N_{\text{tool decisions}}}$ &
$k, S, N_{\text{tools}}$
\\

\textbf{Memory} ($C_M$) &
Ability to retain and retrieve information over a temporal horizon $H$,
$C_M(H)=\Pr(\text{correct recall}\mid H)$ &
$H, L, k$
\\

\textbf{Adaptation} ($C_A$) &
Ability to improve performance following feedback,
$C_A=\frac{C_1-C_0}{1-C_0}$ &
$D_{\text{FT}}, D_{\text{pref}}, C_{\text{PT}}, S$
\\

\textbf{Coordination} ($C_C$) &
Ability to coordinate multiple agents or subtasks toward a common objective,
$C_C=\frac{N_{\text{successful}}}{N_{\text{multi-agent tasks}}}$ &
$N_{\text{agents}}, S, H, T$
\\
\bottomrule
\end{tabularx}
\end{table*}
The capability vector can be defined as in Table \ref{tab:capability_definitions}. These definitions are operational rather than intrinsic. We therefore treat the capability vector as a common coordinate
system for analysing how different scaling regimes affect distinct
functional capabilities, rather than as a prescription for a universal
benchmark for each capability. The resulting capability vector is given by
\begin{equation}
\mathbf{C}
=
\left[
C_K,C_R,C_P,C_T,C_M,C_A,C_C
\right]^{\mathsf T}.
\end{equation}

Each of these capabilities $C_i=f(X)$ naturally are coupled with the pre-, post-, and test-time resources ($X$) as identified in the Table \ref{tab:capability_definitions}. In the following discussion, we will develop mathematical formalism of the capability manifold.
\subsection{Mathematical Formulation}
Let us define the generalised resource vector as
\begin{equation}
\mathbf{X} = [X_1,X_2,...,X_n]^T,
\qquad 
X_i \in \{N, D, C, N_{FT}, N_{Tools},...\}
\end{equation}
The existing scaling-laws where they exists can be converted into bounded scaling-function as follow:
\begin{equation}
\phi_j(X_j) = 1- \Big( \frac{X_{j,o}}{X_{j}} \Big)^{\alpha_j} \textrm{ such that } \phi_j(X_{j,o})=0, \& \lim _{X_j\rightarrow\infty} \phi_j(X_j)=1.
\end{equation}
We can then define an intermediate scaling variable $Z_i$ as
\begin{equation}
Z_i = b_i + \sum_j A_{ij}\phi_j(X_j)+\sum_{j<k} B_{ijk}\phi_j(X_j)\phi_k(X_k)
\end{equation}

where $A_{ij}$ weights relative contribution of resource towards the capability and $B_{ijk}$ being a tensor showing cross-coupling of resources e.g. multiplicative joint power-law as witnessed in post-training. For instance, if $X=[N,D,C]$  and $C=[C_K, C_R, C_P]$ then such cross-coupling is given by:
\begin{equation}
B_i =
\begin{bmatrix}
0 & B_{i,\text{ND}} & B_{i,\text{NC}}\\
B_{i,\text{ND}} & 0 & B_{i,\text{DC}}\\
B_{i,\text{NC}} & B_{i,\text{DC}} & 0
\end{bmatrix}, \quad \text{for } i \in \{K, R, P\}.
\end{equation}
Thus $Z$ can be written as:
\begin{equation}
\mathbf{Z} = \mathbf{b} + \mathbf{A}\phi(\mathbf{X})+\mathcal{B}(\phi(\mathbf{X}),\phi(\mathbf{X})),
\end{equation}
\begin{equation}
\phi(\mathbf{X}) = \begin{bmatrix} 1 - \left(\frac{X_1}{X_{1,0}}\right)^{-\alpha_1} \\ \vdots \\ 1 - \left(\frac{X_n}{X_{n,0}}\right)^{-\alpha_n} \end{bmatrix}
\end{equation}
\begin{equation}
    [\mathcal{B}(\phi(\mathbf{X}),\phi(\mathbf{X}))]_i=\sum_{j<k} B_{ijk}\phi_j(X_j)\phi_k(X_k)
\end{equation}
It is feasible to replace $X_n$ scaling function with a different function $f(X_i)$ when power-law relation does not hold. Now the capability can be defined as:
\begin{equation}
C_i (\boldsymbol{X}) = C_{i,min} + (C_{i,max}-C_{i,min})\sigma(\boldsymbol{Z})
\end{equation}
where $\sigma(x)=1/(1+\exp(-x))$ is standard sigmoid. One can then take Jacobian of the $\mathbf{C}\in \mathbb{R}^m$ as:
\begin{equation}
J_C=\frac{\partial \boldsymbol{C} }{\partial \mathbf{X}} = \underbrace{\frac{\partial \boldsymbol{C} }{\partial \mathbf{Z}}}_{\textrm{saturation}} \underbrace{A+\frac{\partial B}{\partial \phi}}_{\textrm{capability mapping}} \underbrace{\frac{\partial \boldsymbol{\phi} }{\partial \mathbf{X}}}_{\textrm{scaling law}},
\end{equation}
So the Jacobian combines all three key properties of the capability structure, the scaling of capability with the resource can be defined as: 
\begin{equation}
\partial \mathbf{C} =  J_c \partial \mathbf{X},
\end{equation}
Finally, one can define the capability manifold as:
\begin{equation}
\mathcal{M}(\mathbf{C}) = \{\mathbf{C}(\mathbf{X}):\mathbf{X}\in \mathcal{X} \},
\end{equation}
The capability manifold characterises the multidimensional performance of the model as a function of the resources and its architectural attributes.

\begin{figure}[tbp]
    \centering
    \includegraphics[width=\textwidth]{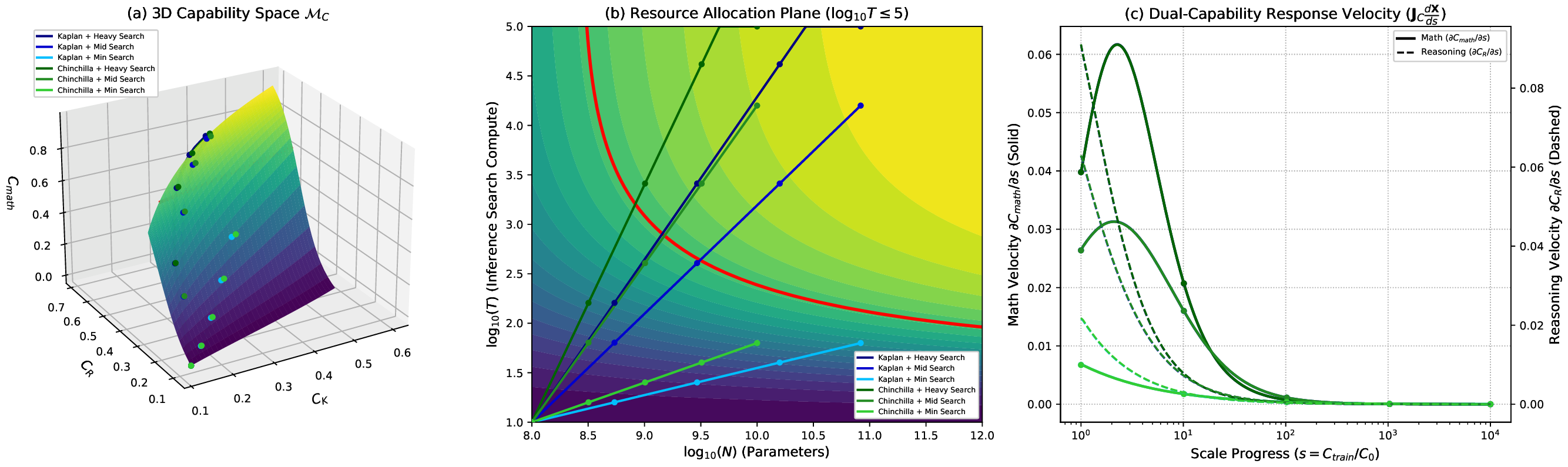}
    \caption{\textbf{Capability Trajectories and Jacobian Velocity Response.} (a) 3D capability space showing emergence thresholds, (b) Resource allocation plane constrained to $\log_{10} T \le 5$, and (c) Dual-axis Jacobian capability velocities ($\partial C_{math}/\partial s$ and $\partial C_{R}/\partial s$) comparing Kaplan vs. Chinchilla scaling strategies across test-time search regimes.}
    \label{fig:capability_trajectories}
\end{figure}

\subsection{Application of Framework}
\begin{figure}[tbp]
    \centering
    \includegraphics[width=0.5\textwidth]{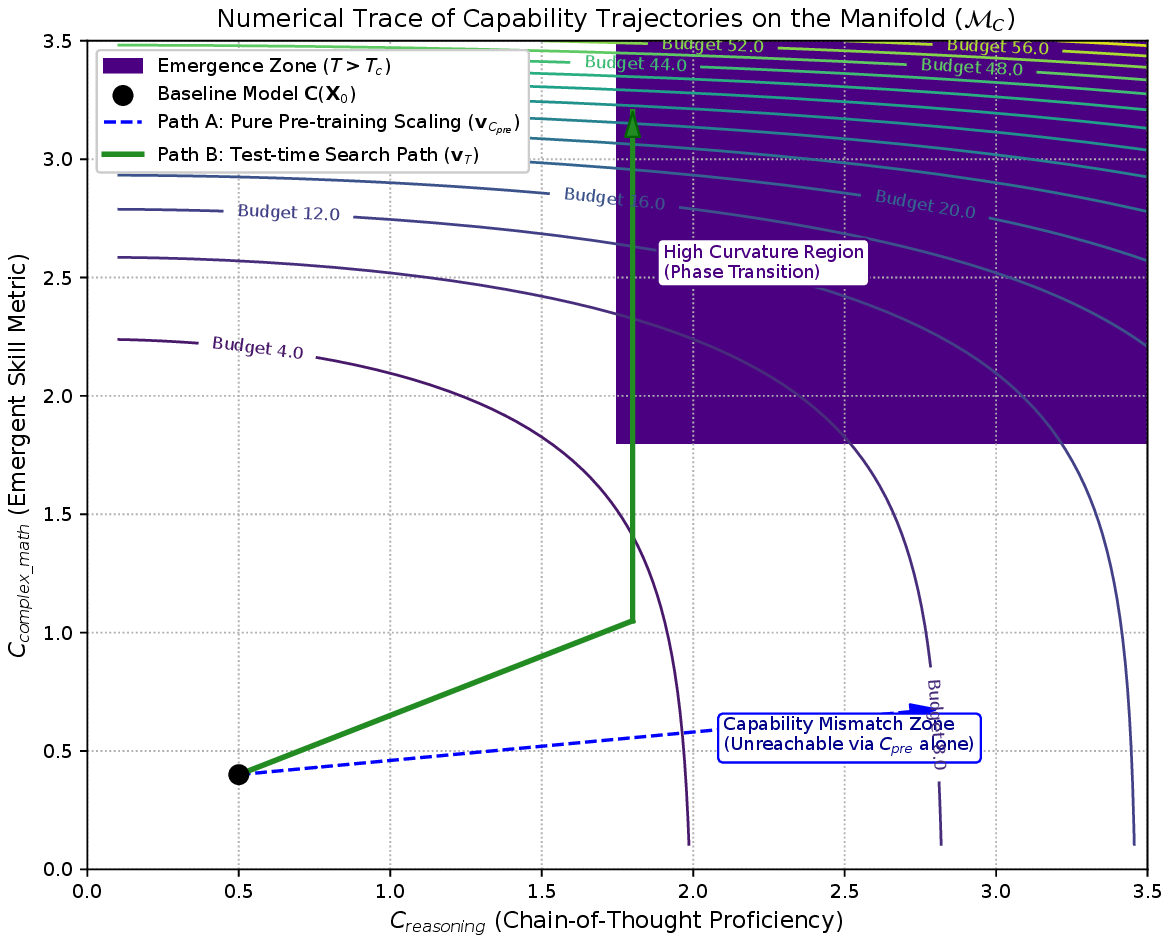}
    \caption{\textbf{Numerical Trace of Capability Trajectories on the Manifold $\mathcal{M}_C$.} Iso-budget contours show compute requirements as a function of $C_{\mathrm{reasoning}}$ and  $C_{\mathrm{complex\_math}}$. Path~A represents pure pre-training scaling ($\mathbf{v}_{C_{\mathrm{pre}}}$), which leads to a capability mismatch zone. Path~B demonstrates test-time search scaling ($\mathbf{v}_T$), triggering a phase transition ($T > T_c$).}
    \label{fig:percolation}
\end{figure}

To demonstrate how this framework unifies resource scaling laws into capability dynamics, consider $\mathbf{X} = [N, D, T]^\top$ incorporating test-time search $T(s) = \min(T_0 \cdot s^{T_{\mathrm{mult}}}, 10^5)$ mapped to downstream capabilities $\mathbf{C} = [C_K, C_R, C_M]^\top$, deriving analytical Jacobians $\mathbf{J}_C = \frac{\partial \mathbf{C}}{\partial \mathbf{X}}$ across Kaplan and Chinchilla paradigms paired with varying search intensities $T_{\mathrm{mult}} \in \{1.20, 0.80, 0.20\}$. Trajectory analysis (Fig. ~\ref{fig:capability_trajectories}) reveals a decoupling between training and inference scaling: while pre-training allocation dictates baseline reasoning velocity ($\partial C_R / \partial s$), emergent math velocity ($\partial C_{\mathrm{math}} / \partial s$) is governed by $T_{\mathrm{mult}}$, enabling search-heavy models to achieve steeper capability acceleration while pre-training-heavy paths plateau. Topological mapping on the manifold $\mathcal{M}_C$ under an activation energy model $B = C_{\mathrm{reasoning}}^2 + 0.5 \exp(2.0(C_{\mathrm{complex\_math}} - 1.2))$ confirms that pure pre-training (Path~A) drifts into a capability mismatch zone, whereas test-time search scaling (Path~B) triggers a sharp phase transition near $T > T_c = 1.8$ to overcome exponential energy barriers (Fig.~\ref{fig:percolation}).

\subsection{Conclusion}
In this paper, we presented the capability manifold, a unified framework mapping heterogeneous scaling resources to trajectories in a multidimensional capability space rather than a single loss metric. By integrating pre-training, post-training, and test-time compute, this framing reveals capability-dependent marginal returns and enables direct comparison across alternative compute strategies. Ultimately, it provides a basis for analysing efficient scaling, distillation, safety, and capability trade-offs in foundation models.
\bibliographystyle{plainnat} 
\bibliography{references}

\newpage
\section*{NeurIPS Paper Checklist}

\begin{enumerate}

\item {\bf Claims}
    \item[] Question: Do the main claims made in the abstract and introduction accurately reflect the paper's contributions and scope?
    \item[] Answer: \answerYes{} 
    
\item {\bf Limitations}
    \item[] Question: Does the paper discuss the limitations of the work performed by the authors?
    \item[] Answer: \answerYes{} 

\item {\bf Theory assumptions and proofs}
    \item[] Question: For each theoretical result, does the paper provide the full set of assumptions and a complete (and correct) proof?
    \item[] Answer: \answerYes{} 
    
    \item {\bf Experimental result reproducibility}
    \item[] Question: Does the paper fully disclose all the information needed to reproduce the main experimental results of the paper to the extent that it affects the main claims and/or conclusions of the paper (regardless of whether the code and data are provided or not)?
    \item[] Answer: \answerYes{} 

\item {\bf Open access to data and code}
    \item[] Question: Does the paper provide open access to the data and code, with sufficient instructions to faithfully reproduce the main experimental results, as described in supplemental material?
    \item[] Answer: \answerYes{} 
    \item[] Justification: The results are produced from simulation of mathematical equations and all parameters are given in the paper.
    
\item {\bf Experimental setting/details}
    \item[] Question: Does the paper specify all the training and test details (e.g., data splits, hyperparameters, how they were chosen, type of optimizer) necessary to understand the results?
    \item[] Answer: \answerYes{} 

\item {\bf Experiment statistical significance}
    \item[] Question: Does the paper report error bars suitably and correctly defined or other appropriate information about the statistical significance of the experiments?
    \item[] Answer: \answerYes{} 
    \item[] Justification: All simulation plots include shaded error bands representing standard deviation across independent random seeds.
    
\item {\bf Experiments compute resources}
    \item[] Question: For each experiment, does the paper provide sufficient information on the computer resources (type of compute workers, memory, time of execution) needed to reproduce the experiments?
    \item[] Answer: \answerYes{} 
    \item[] Justification: Compute specifications, hardware details, and runtimes are detailed in Appendix Section B.
    
\item {\bf Code of ethics}
    \item[] Question: Does the research conducted in the paper conform, in every respect, with the NeurIPS Code of Ethics \url{https://neurips.cc/public/EthicsGuidelines}?
    \item[] Answer: \answerYes{} 
    \item[] Justification: The research adheres entirely to the NeurIPS Code of Ethics.

\item {\bf Broader impacts}
    \item[] Question: Does the paper discuss both potential positive societal impacts and negative societal impacts of the work performed?
    \item[] Answer: \answerYes{} 
    \item[] Justification: Potential broader impacts and broader implications are discussed in Section 6.
    
\item {\bf Safeguards}
    \item[] Question: Does the paper describe safeguards that have been put in place for responsible release of data or models that have a high risk for misuse (e.g., pre-trained language models, image generators, or scraped datasets)?
    \item[] Answer: \answerNA{} 
    \item[] Justification: This work does not release high-risk models or sensitive datasets.

\item {\bf Licenses for existing assets}
    \item[] Question: Are the creators or original owners of assets (e.g., code, data, models), used in the paper, properly credited and are the license and terms of use explicitly mentioned and properly respected?
    \item[] Answer: \answerYes{} 
    \item[] Justification: All software packages and baseline code frameworks used are explicitly cited with their standard open-source licenses noted in the appendix.

\item {\bf New assets}
    \item[] Question: Are new assets introduced in the paper well documented and is the documentation provided alongside the assets?
    \item[] Answer: \answerNA{} 
    \item[] Justification: No new dataset or standalone software library was introduced.

\item {\bf Crowdsourcing and research with human subjects}
    \item[] Question: For crowdsourcing experiments and research with human subjects, does the paper include the full text of instructions given to participants and screenshots, if applicable, as well as details about compensation (if any)? 
    \item[] Answer: \answerNA{} 
    \item[] Justification: This paper does not involve crowdsourcing or human subjects.

\item {\bf Institutional review board (IRB) approvals or equivalent for research with human subjects}
    \item[] Question: Does the paper describe potential risks incurred by study participants, whether such risks were disclosed to the subjects, and whether Institutional Review Board (IRB) approvals (or an equivalent approval/review based on the requirements of your country or institution) were obtained?
    \item[] Answer: \answerNA{} 
    \item[] Justification: This paper does not involve human subjects research.

\item {\bf Declaration of LLM usage}
    \item[] Question: Does the paper describe the usage of LLMs if it is an important, original, or non-standard component of the core methods in this research? Note that if the LLM is used only for writing, editing, or formatting purposes and does \emph{not} impact the core methodology, scientific rigor, or originality of the research, declaration is not required.
    \item[] Answer: \answerNA{} 
    \item[] Justification: LLMs are not a core component of the proposed methodology or research pipeline.

\end{enumerate}

\end{document}